# Discovering Classes of Strongly Equivalent Logic Programs

**Fangzhen Lin**　　　　　　　　　　　　　　　　　　　　　　　　　FLIN@CS.UST.HK
*Department of Computer Science and Engineering*
*Hong Kong University of Science and Technology*
*Clear Water Bay, Kowloon, Hong Kong*

**Yin Chen**　　　　　　　　　　　　　　　　　　　　　　　　　GZCHENYIN@GMAIL.COM
*Department of Computer Science*
*South China Normal University*
*Guangzhou, P.R. China*

## Abstract

In this paper we apply computer-aided theorem discovery technique to discover theorems about strongly equivalent logic programs under the answer set semantics. Our discovered theorems capture new classes of strongly equivalent logic programs that can lead to new program simplification rules that preserve strong equivalence. Specifically, with the help of computers, we discovered exact conditions that capture the strong equivalence between a rule and the empty set, between two rules, between two rules and one of the two rules, between two rules and another rule, and between three rules and two of the three rules.

## 1. Introduction

In this paper we apply computer-aided theorem discovery technique to discover theorems about strongly equivalent logic programs under the answer set semantics. Our discovered theorems capture new classes of strongly equivalent logic programs that can lead to new program simplification rules that preserve strong equivalence.

Theorem discovery is a highly creative human process. Generally speaking, we can divide it into two steps: (i) conjecture formulation, and (ii) conjecture verification, and computers can help in both of these two steps. For instance, machine learning tools can be used in the first step, i.e. in coming up with reasonable conjectures, and automated deduction tools can be used in the second step, i.e. in verifying the correctness of these conjectures.

While theorem discovery may make use of learning, these two tasks are fundamentally different. Theorem discovery starts with a theory, and aims at finding *interesting* consequences of the theory, while learning is mostly about induction, i.e. it starts with examples/consequences, and aims at finding a theory that would explain the given examples/consequences.

Using computers to discover theorems is an old aspiration. There have been some success stories. For instance, AM (Lenat, 1979) was reported to be able to come up with some interesting concepts and theorems in number theory, and the remarkable systems described by Petkovsek, Wilf, and Zeilberger (1996) can discover many identities, especially hypergeometric identities involving sums of binomial coefficients that are important for the analyses of algorithms. Yet another example where "interesting" theorems can be discovered almost fully automatically is a recent work by Lin (2004) on discovering state invariants in planning domains. Lin showed that there are ways to classify many state constraints that are useful in planning according to their syntactic properties, and enumerate them easily for many domains. Furthermore, for many of these constraints whether they are invariants can be checked automatically. As a result, the system described by Lin (2004) can discover many common constraints in planning domains, and for the logistics domain, it could even discover a set of "complete" state invariants.

Following this line of research, in this paper, we consider the problem of discovering classes of strongly equivalent sets of logic program rules under answer set semantics. As noted by Lifschitz, Pearce, and Valverde (2001), if two sets of rules are strongly equivalent, then we can replace one by the other in any logic program without changing the semantics of the program. Thus identifying strongly equivalent sets of logic program rules is a useful exercise that may have applications in program simplification.

This paper is organized as follows. In the next section, we briefly review the basic concepts of logic programming under answer set semantics. Then in section 3 we state in more precise terms the type of theorems that we want to discover. In section 4 we prove some general theorems that will help us prove these theorems, and in section 5, we describe some of the theorems that we discovered. We then discuss an application to logic program simplification in section 6, and finally we conclude this paper in section 7.

## 2. Answer Set Programming

Traditional logic programming systems like Prolog solve problems by query answering. The user encodes knowledge about a domain by a set of rules, and solves a problem by issuing queries to the set of rules. In contrast, Answer Set Programming (ASP) (Niemelä, 1999; Lifschitz, 1999; Marek & Truszczynski, 1999) is a constraint-based programming paradigm. It is based on logic programming with answer set semantics (Gelfond & Lifschitz, 1988, 1991). To solve a problem, the user encodes the domain knowledge as a logic program in such a way that the answer sets of the program will correspond to the solutions to the original problem. Compared to other constraint-based programming paradigms, ASP allows natural encodings of recursive relations, and has built-in facilities for default reasoning. Several ASP solvers have been developed (Niemelä, Simons, & Syrjänen, 2000; Leone, Pfeifer, Faber, Eiter, Gottlob, Perri, & Scarcello, 2006; Lin & Zhao, 2004; Lierler & Maratea, 2004). To date, ASP has been used in space shuttle planning (Nogueira, Balduccini, Gelfond, Watson, & Barry, 2001), evolutional linguistics (Erdem, Lifschitz, Nakhleh, & Ringe, 2003), and others. In the following, we briefly review some basic notions in ASP.

Let $L$ be a propositional language, i.e. a set of atoms. In this paper we shall consider logic programs with rules of the following form:

$$h_1; \cdots ; h_k \leftarrow p_1, \cdots, p_m, not\, p_{m+1}, \cdots, not\, p_n \qquad (1)$$

where $h_i$'s and $p_i$'s are atoms in $L$. So a logic program here can have default negation (*not*), constraints (when $k = 0$), and disjunctions in the head of its rules. In the following, if $r$ is a rule of the above form, we write $Hd_r$ to denote the set $\{h_1, ..., h_k\}$, $Ps_r$ the set $\{p_1, ..., p_m\}$, and $Ng_r$ the set $\{p_{m+1}, ..., p_n\}$. Thus a rule $r$ can also be written as $Hd_r \leftarrow Ps_r, not\, Ng_r$. The semantics of these programs are given by *answer sets* (Gelfond & Lifschitz, 1991), which are defined by a fixed-point operator through what has been known as *Gelfond-Lifschitz transformation*. Let $X$ be a subset of $L$, and $P$ a logic program. The Gelfond-Lifschitz transformation of $P$ on $X$, written $P^X$, is the set of rules obtained from $P$ according to the following two rules:

1. If a rule of the form (1) is in $P$, and $p_i \in X$ for some $m + 1 \leq i \leq n$, then delete this rule.

2. Delete all literals of the form $not\, p_i$ in the bodies of the remaining rules.

For instance, if $P$ is the set of following rules:

$$a; b \leftarrow$$
$$c \leftarrow not\, a$$

then $P^{\{a\}}$ is $\{a; b \leftarrow\}$, and $P^{\{b\}}$ is $\{(a; b \leftarrow), (c \leftarrow)\}$.

Clearly, for any $X$ and $P$, $P^X$ is a set of rules which do not have the "*not*" operator. Now a set $X$ is an answer set of $P$ if $X$ is a minimal set of atoms that *satisfies* every rule in $P^X$, where $X$ satisfies a rule of the form

$$h_1; \cdots; h_k \leftarrow p_1, \cdots, p_m$$

if for some $1 \leq i \leq k$, $h_i \in X$ whenever $\{p_1, ..., p_m\} \subseteq X$. For instance, for the above program, both $\{a\}$ and $\{b, c\}$ are answer sets, and they are the only answer sets of the program.

Two logic programs $P_1$ and $P_2$ are said to be *equivalent* if they have the same answer sets, and *strongly equivalent* (Lifschitz et al., 2001) (under the language $L$), written $P_1 \simeq_{se} P_2$, if for any logic program $P$ in $L$, $P \cup P_1$ and $P \cup P_2$ are equivalent (thus we write $P_1 \not\simeq_{se} P_2$ when $P_1$ and $P_2$ are not strongly equivalent). For example, $\{a \leftarrow b\}$ and $\{a \leftarrow c\}$ are equivalent, but not strongly equivalent. It can be shown that $\{a \leftarrow not\, a\} \simeq_{se} \{\leftarrow not\, a\}$. As in the abstract, we also say that a rule $r$ is strongly equivalent to another rule $r'$, written $r \simeq_{se} r'$, if $\{r\} \simeq_{se} \{r'\}$, and two rules $r_1$ and $r_2$ are strongly equivalent to a rule $r$, written $\{r_1, r_2\} \simeq_{se} r$, if $\{r_1, r_2\} \simeq_{se} \{r\}$, and so on.

The notion of strong equivalence is important for ASP for several reasons. First of all, it helps us understand the answer set semantics. For instance, Turner (2003) showed that the disjunctive rule $(a; b \leftarrow)$ is not strongly equivalent to any set of normal rules. This implies that there cannot be a modular translation from disjunctive logic programs to normal logic programs. However,

$$\{(a; b \leftarrow), (\leftarrow a, b)\}$$

is strongly equivalent to

$$\{(a \leftarrow not\, b), (b \leftarrow not\, a), (\leftarrow a, b)\}.$$

This means that under the constraint $(\leftarrow a, b)$, the disjunctive rule $(a; b \leftarrow)$ can be replaced by two rules without disjunction. Secondly, as we mentioned in the introduction, if $P_1$ and $P_2$ are strongly equivalent, then they are interchangeable regardless of where they occur. Thus if we have a large repertoire of pairs of strongly equivalent logic programs, we could use them to transform a given program into one that is most suitable to the need in hand. In particular, it could help us simplify a program for the purpose of computing its answer sets. As we shall see, our discovered theorems will contribute significantly to this repertoire.

Lifschitz et al. (2001) showed that checking for strong equivalence between two logic programs can be done in the logic of here-and-there, a three-valued non-classical logic somewhere between classical logic and intuitionistic logic. Lin (2002) provided a mapping from logic programs to propositional theories and showed that two logic programs are strongly equivalent iff their corresponding theories in propositional logic are equivalent. This result will be used here both for generating example pairs of strongly equivalent logic programs, and for verifying a conjecture. We repeat it here.

Let $P_1$ and $P_2$ be two finite logic programs, and $L$ the set of atoms in them.

**Theorem 1** *(Lin, 2002)* $P_1 \simeq_{se} P_2$ *iff in propositional logic, the following sentence is valid:*

$$(\bigwedge_{p \in L} p \supset p') \supset [\bigwedge_{r \in P_1} \delta(r) \equiv \bigwedge_{r \in P_2} \delta(r)], \tag{2}$$

*where for each $p \in L$, $p'$ is a new atom, and for each rule $r$ of the form (1), $\delta(r)$ is the conjunction of the following two sentences:*

$$p_1 \wedge \cdots \wedge p_m \wedge \neg p'_{m+1} \wedge \cdots \wedge \neg p'_n \supset h_1 \vee \cdots \vee h_k, \tag{3}$$

$$p'_1 \wedge \cdots \wedge p'_m \wedge \neg p'_{m+1} \wedge \cdots \wedge \neg p'_n \supset h'_1 \vee \cdots \vee h'_k. \tag{4}$$

Notice that if $m = n = 0$, then the left sides of the implications in (3) and (4) are considered to be *true*, and if $k = 0$, then the right sides of the implications in (3) and (4) are considered to be *false*.

In general checking if two sets of rules are strongly equivalent is coNP-complete (c.f. Turner, 2001; Pearce, Tompits, & Woltran, 2001; Lin, 2002).

## 3. The Problem

As we mentioned above, one possible use of the notion of strongly equivalent logic programs is in program simplification. For instance, given a logic program, for each rule $r$ in it, we may ask whether it can be deleted without knowing what other rules are in $P$, i.e. whether $\{r\}$ is strongly equivalent to the empty set. Or we may ask whether a rule $r$ in $P$ can be deleted if one knows that another rule $r'$ is already in $P$, i.e. whether $\{r, r'\}$ is strongly equivalent to $\{r'\}$. In general, we may ask the following *k-m-n* question: Is $\{r_1, ..., r_k, u_1, ..., u_m\} \simeq_{se} \{r_1, ..., r_k, v_1, ..., v_n\}$? Thus our theorem discovery task is to come up, for a given *k-m-n* problem, a computationally effective condition that holds if and only if the answer to the *k-m-n* question is positive.

Now suppose we have such a condition $C$, and suppose that when

$$\{r_1, ..., r_k, u_1, ..., u_m\} \simeq_{se} \{r_1, ..., r_k, v_1, ..., v_n\},$$

it is better to replace $\{u_1, ..., u_m\}$ by $\{v_1, ..., v_n\}$ in the presence of $r_1, ..., r_k$ for the purpose of, say computing the answer sets of a program. One way to use this result to simplify a given program $P$ is to first choose $k$ rules in $P$, and for any other $m$ rules in it, try to find $n$ rules so that the condition $C$ holds, and then replace the $m$ rules in $P$ by the simpler $n$ rules.

However, even if checking whether $C$ holds would take a negligible constant time, using the above procedure to simplify a given logic program will be practical only when $k, m, n$ are all very small or when $k$ is almost the same as the number of the rules in the given program, and $m$ and $n$ are very small. Thus it seems to us that it is worthwhile to solve the $k$-$m$-$n$ problem only when $k, m, n$ are small. In particular, in this paper, we shall concentrate on the 0-1-0 problem (whether a rule can always be deleted), the 0-1-1 problem (whether a rule can always be replaced by another one), the 1-1-0 problem (in the presence of a rule, whether another rule can be deleted), the 2-1-0 problem (in the presence of two rules, whether a rule can always be deleted), and the 0-2-1 problem (if a pair of rules can be replaced by a single rule).

An example of theorems that we want to discover about these problems is as follows:

$$\text{For any rule } r, \; r \simeq_{se} \emptyset \text{ iff } (Hd_r \cup Ng_r) \cap Ps_r \neq \emptyset. \tag{5}$$

## 4. Some General Theorems

In this section, we prove some general theorems that will help us verify whether an assertion like (5) above is true.

Let $L$ be a propositional language, i.e. a set of atoms. From $L$, construct a first-order language $F_L$ with equality, two unary predicates $H_1$ and $H_2$, three unary predicates $Hd_r$, $Ps_r$, and $Ng_r$ for each logic program rule $r$ in $L$ (we assume that each rule in $L$ has a unique name), and three unary predicates $X_i$, $Y_i$, and $Z_i$ for each positive number $i$.

Notice that we have used $Hd_r$, $Ps_r$, and $Ng_r$ to denote sets of atoms previously, but now we overload them as unary predicates. Naturally, the intended interpretations of these unary predicates are their respective sets.

**Definition 1** *Given a set $L$ of atoms, an* intended model *of $F_L$ is one whose domain is $L$, and for each rule $r$ in $L$, the unary predicates $Ps_r$, $Hd_r$, and $Ng_r$ are interpreted by their corresponding sets of atoms, $Ps_r$, $Hd_r$, and $Ng_r$, respectively.*

Conditions on rules in $L$, such as $Ps_r \cap Ng_r \neq \emptyset$, will be expressed by special sentences called *properties* in $F_L$.

**Definition 2** *A sentence of $F_L$ is a* property about $n$ rules *if it is constructed from equality and predicates $X_i$, $Y_i$, and $Z_i$, $1 \leq i \leq n$. A property $\Phi$ about $n$ rules is true (holds) on a sequence $P = [r_1, ..., r_n]$ of $n$ rules if $\Phi[P]$ is true in an intended model of $F_L$, where $\Phi[P]$ is obtained from $\Phi$ by replacing each $X_i$ by $Hd_{r_i}$, $Y_i$ by $Ps_{r_i}$, and $Z_i$ by $Ng_{r_i}$.*

Notice that since $\Phi[P]$ does not mention predicates $X_i$, $Y_i$, $Z_i$, $H_1$, and $H_2$, if it is true in one intended model, then it is true in all intended models.

As we have mentioned above, we are interested in capturing the strong equivalence between two programs by a computationally effective condition. More specifically, for some small $k$, $m$, and $n$, we are interested in finding a property $\Phi$ about $k + m + n$ rules such that for any sequence of $k + m + n$ rules, $P = [r_1, ..., r_k, u_1, ..., u_m, v_1, ..., v_n]$,

$$\{r_1, ..., r_k, u_1, ..., u_m\} \simeq_{se} \{r_1, ..., r_k, v_1, ..., v_n\} \text{ iff } \Phi \text{ is true on } P. \tag{6}$$

We shall now prove some general theorems that can help us verify the above assertion for a class of formulas $\Phi$.

First of all, Theorem 1 can be reformulated in $F_L$ as follows by reading $H_1(p)$ as "$p$ holds", and $H_2(p)$ as "$p'$ holds":

**Theorem 2** $P_1 \simeq_{se} P_2$ in $L$ iff the following sentence

$$\forall x(H_1(x) \supset H_2(x)) \supset [\bigwedge_{r \in P_1} \gamma(r) \equiv \bigwedge_{r \in P_2} \gamma(r)] \tag{7}$$

is true in all intended models of $F_L$, where $\gamma(r)$ is the conjunction of the following two sentences:

$$[\forall x(Ps_r(x) \supset H_1(x)) \wedge \forall x(Ng_r(x) \supset \neg H_2(x))] \supset \exists x(Hd_r(x) \wedge H_1(x)), \tag{8}$$
$$[\forall x(Ps_r(x) \supset H_2(x)) \wedge \forall x(Ng_r(x) \supset \neg H_2(x))] \supset \exists x(Hd_r(x) \wedge H_2(x)). \tag{9}$$

In first order logic, if a prenex formula of the form $\exists \vec{x} \forall \vec{y} B$ is satisfiable, then it is satisfiable in a structure with $n$ elements, where $B$ is a formula that contains no quantifiers, constants, or function symbols, and $n$ is the length of $\vec{x}$ if it is non-empty, and 1 when $\vec{x}$ is empty. We can prove a similar result for our first-order languages and their intended models here.

**Definition 3** *A sentence of $F_L$ is an* extended *property about $n$ rules if it is constructed from equality and predicates $X_i$, $Y_i$, and $Z_i$, $1 \le i \le n$, and $H_1$ and $H_2$. An extended property $\Phi$ about $n$ rules is true (holds) on a sequence $P = [r_1, ..., r_n]$ of $n$ rules in a model $M$ if $\Phi[P]$ is true in $M$, where $\Phi[P]$ is obtained from $\Phi$ by replacing each $X_i$ by $Hd_{r_i}$, $Y_i$ by $Ps_{r_i}$, and $Z_i$ by $Ng_{r_i}$.*

**Definition 4** *In the following, if $P = [r_1, ..., r_n]$ is a tuple of rules in $L$, and $L'$ is a subset of $L$, then we define the* restriction *of $P$ on $L'$ to be $[r'_1, ..., r'_n]$, where $r'_i$ is*

$$Hd_{r_i} \cap L' \leftarrow Ps_{r_i} \cap L', not\,(Ng_{r_i} \cap L').$$

**Lemma 1** *Let $\Phi$ be an extended property in $F_L$ about $n$ rules, and of the form $\exists \vec{x} \forall \vec{y} Q$, where $\vec{x}$ is a tuple of $w$ variables, and $Q$ a formula that does not have any quantifiers. If $\Phi$ holds on a sequence $P$ of $n$ rules in an intended model $M$ of $F_L$, then there is a subset $L'$ of $L$ such that $L'$ has at most $w$ atoms (or one atom when $w = 0$), and $\Phi$ holds on the restriction of $P$ on $L'$ in an intended model of $F_{L'}$.*

**Proof:** Suppose $M$ is an intended model of $F_L$ such that $M \models \Phi[P]$. Thus there is a tuple $\vec{p}$ of $w$ (or one when $w = 0$) atoms in $L$ such that $M \models \forall \vec{y} Q[P](\vec{x}/\vec{p})$. Now let $L'$ be the set of atoms in $\vec{p}$, and $M'$ defined as follows:

- Each of the predicates $H_1$, $H_2$, $X_i$, $Y_i$, and $Z_i$, $i \ge 1$, is interpreted as the restriction of its interpretation in $M$ on $L'$.

- For each rule $r$ in $L'$, the predicates $Hd_r$, $Ps_r$, and $Ng_r$ are interpreted the same as they are in $M$. This is well-defined as $r$ is also a rule in $L$,

Then $M'$ is an intended model of $F_{L'}$. Let $P'$ be the restriction of $P$ on $L'$. Then $P'$ is a tuple of rules in $L'$. Since $Q$ has no quantifiers (and the language has no function symbols), for any instantiation $\vec{u}$ of $\vec{y}$ in $L'$, $M \models Q[P](\vec{x}/\vec{p})(\vec{y}/\vec{u})$ iff $M' \models Q[P'](\vec{x}/\vec{p})(\vec{y}/\vec{u})$. Since $M \models \forall \vec{y} Q[P](\vec{x}/\vec{p})$, we have $M' \models \forall \vec{y} Q[P'](\vec{x}/\vec{p})$, Thus $M' \models \exists \vec{x} \forall \vec{y} Q[P']$. $\square$

Using Theorem 2 and this lemma, we can show the following theorem which will enable us to automate the verification of the "if" part of (6) when the property $\Phi$ is in the prenex format.

**Theorem 3** *Without loss of generality, suppose $m \geq n$. If $\Phi$ is a property about $k+m+n$ rules of the form $\exists \vec{x} \forall \vec{y} Q$, where $\vec{x}$ is a tuple of $w$ variables, and $Q$ a formula that does not have any quantifiers, then the following two assertions are equivalent:*

(a) *For any sequence of $k + m + n$ rules, $P = [r_1, ..., r_k, u_1, ..., u_m, v_1, ..., v_n]$, if $\Phi$ is true on $P$, then $\{r_1, ..., r_k, u_1, ..., u_m\} \simeq_{se} \{r_1, ..., r_k, v_1, ..., v_n\}$.*

(b) (b.1) *If $n > 0$, then for any sequence $P = [r_1, ..., r_k, u_1, ..., u_m, v_1, ..., v_n]$ of rules with at most $w + 2(k + m)$ atoms, if $\Phi$ is true on $P$, then*

$$\{r_1, ..., r_k, u_1, ..., u_m\} \simeq_{se} \{r_1, ..., r_k, v_1, ..., v_n\}.$$

(b.2) *If $n = 0$, then for any sequence $P = [r_1, ..., r_k, u_1, ..., u_m]$ of rules with at most $K$ atoms, if $\Phi$ is true on $P$, then*

$$\{r_1, ..., r_k, u_1, ..., u_m\} \simeq_{se} \{r_1, ..., r_k\},$$

*where $K$ is $w + 2k$ if $w + 2k > 0$, and $K = 1$ otherwise.*

**Proof:** If (a) then (b) is obvious. We assume that (b) is true, and show that (a) holds as well. Suppose first that $n > 0$. Suppose $P = [r_1, ..., r_k, u_1, ..., u_m, v_1, ..., v_n]$ is a sequence of $k + m + n$ rules in a language $L$ such that $\Phi$ is true on $P$, and

$$\{r_1, ..., r_k, u_1, ..., u_m\} \not\simeq_{se} \{r_1, ..., r_k, v_1, ..., v_n\}.$$

Thus there is an intended model of $F_L$ that satisfies $\Phi[P]$, and an intended model $M$ of $F_L$ that satisfies the following sentence:

$$(\forall x) H_1(x) \supset H_2(x) \wedge \neg [\bigwedge_{r \in P_1} \gamma(r) \equiv \bigwedge_{r \in P_2} \gamma(r)],$$

where $P_1 = \{r_1, ..., r_k, u_1, ..., u_m\}$, and $P_2 = \{r_1, ..., r_k, v_1, ..., v_n\}$. As we noted after Definition 2, $M$ will also satisfy $\Phi[P]$. Thus $M$ satisfies the following sentence

$$\Phi[P] \wedge (\forall x) H_1(x) \supset H_2(x) \wedge \{ [\bigwedge_{r \in P_1} \gamma(r) \wedge \neg \bigwedge_{r \in P_3} \gamma(r)] \vee [\bigwedge_{r \in P_2} \gamma(r) \wedge \neg \bigwedge_{r \in P_4} \gamma(r)] \}, \quad (10)$$

where $P_3 = \{v_1, ..., v_n\}$, and $P_4 = \{u_1, ..., u_m\}$.

Now for any rule $r$, there is an extended property $\varphi(x, y)$ of one rule that does not mention any quantifiers such that $\gamma(r)$ is equivalent to $\exists x, y. \varphi[r]$. Thus for any tuple $Q$ of $t$ rules, there is an extended property $\varphi$ of $t$ rules that does not mention any quantifiers such that $\bigwedge_{r \in Q} \gamma(r)$ is equivalent to $\exists \vec{y}. \varphi[Q]$, where $\vec{y}$ is a tuple of $2t$ variables.

Thus there is

- a tuple $\vec{z_1}$ of $2(k+m)$ variables, a tuple $\vec{z_2}$ of variables, an extended property $\varphi_1$ of $k+m+n$ rules that does not have any quantifiers, and whose free variables are in $\vec{z_1}$ and $\vec{z_2}$; and

- a tuple $\vec{z_3}$ of $2(k+n)$ variables, a tuple $\vec{v_4}$ of variables, and an extended property $\varphi_2$ of $k+m+n$ rules that does not have any quantifiers, and whose free variables are in $\vec{z_3}$ and $\vec{z_4}$ such that $\vec{v_1}, \vec{v_2}, \vec{v_3}, \vec{v_4}$ do not have common variables in them, and (10) is equivalent to the following sentence:

$$\{\Phi \wedge \forall x(H_1(x) \supset H_2(x)) \wedge (\exists \vec{z_1} \forall \vec{z_2} \varphi_1 \vee \exists \vec{z_3} \forall \vec{z_4} \varphi_2)\}[P].$$

Since we have assumed that $m \geq n$, thus there is an extended property $\varphi_3$ about $k+m+n$ rules that does not mention any quantifiers and function symbols, and whose free variables are among $\vec{z_1}$, $\vec{z_2}$, and $\vec{z_4}$ such that the above sentence is equivalent to the following sentence:

$$(\Phi \wedge \forall x(H_1(x) \supset H_2(x)) \wedge \exists \vec{z_1}(\forall \vec{z_2}, \vec{z_4})\varphi_3)[P].$$

Now given the form of $\Phi$ assumed in the theorem, there is a tuple $\vec{z_5}$ of $w + 2(k+m)$ variables, a tuple $\vec{z_6}$ of variables, and an extended property $\Psi$ of $k+m+n$ rules that does not mention any quantifiers, and whose free variables are among $\vec{z_5}$, and $\vec{z_6}$ such that the above sentence is equivalent to $(\exists \vec{z_5})(\forall \vec{z_6})\Psi[P]$.

By Lemma 1, there is a subset $L'$ of $L$ that has at most $w + 2(k+m)$ atoms such that $(\exists \vec{z_5})(\forall \vec{z_6})\Psi$ holds on $P'$, where $P'$ is the restriction of $P$ on $L'$. If

$$P' = [r'_1, ..., r'_k, u'_1, ..., u'_m, v'_1, ..., v'_n],$$

then this will mean that $\Phi$ is true on $P'$, and $\{r'_1, ..., r'_k, u'_1, ..., u'_m\} \not\equiv_{se} \{r'_1, ..., r'_k, v'_1, ..., v'_n\}$. This shows that if (b.1), then (a).

The proof that if (b.2) then (a) is exactly the same except now that

$$[\bigwedge_{r \in P_1} \gamma(r) \equiv \bigwedge_{r \in P_2} \gamma(r)]$$

is equivalent to

$$[\bigwedge_{r \in P_2} \gamma(r) \supset \bigwedge_{r \in P_1} \gamma(r)].$$

□

The "only if" part of (6) can often be proved with the help of the following theorem.

**Theorem 4** *Let $L_1$ and $L_2$ be two languages, and $f$ a function from $L_1$ to $L_2$. If $P_1$ and $P_2$ are two programs in $L_1$ that are strongly equivalent, then $f(P_1)$ and $f(P_2)$ are two programs in $L_2$ that are also strongly equivalent. Here $f(P)$ is obtained from $P$ by replacing each atom $p$ in it by $f(p)$.*

**Proof:** By Theorem 1 and the fact that in propositional logic, if $\varphi$ is a tautology, and $f$ a function from $L_1$ to $L_2$, then $f(\varphi)$ is also a tautology, where $f(\varphi)$ is the formula obtained from $\varphi$ by replacing each atom $p$ in it by $f(p)$. □

For an example of using the theorems in this section for proving assertions of the form (6), see Section 5.1.

## 5. Computer-Aided Theorem Discovery

Given a *k-m-n* problem, our strategy for discovering theorems about it is as follows:

1. Choose a small language L;

2. Generate all possible triples

$$(\{r_1, ..., r_k\}, \{u_1, ..., u_m\}, \{v_1, ..., v_n\}) \qquad (11)$$

   of sets of rules in $L$ such that $\{r_1, ..., r_k, u_1, ..., u_m\} \simeq_{se} \{r_1, ..., r_k, v_1, ..., v_n\}$ in $L$;

3. Formulate a conjecture on the *k-m-n* problem that holds in the language $L$, i.e. a condition that is true for a triple of the form (11) iff it is generated in Step 2;

4. Verify the correctness of this conjecture in the general case.

This process may have to be iterated. For instance, a conjecture formulated in Step 3 may fail to generalize in Step 4, so we either need to formulate a new conjecture or start all over again in step 1 using a larger language.

Ideally, we would like this process to be automatic. However, it is difficult to automate Steps 3 and 4 - the number of possible patterns that we need to examine in order to come up with a good conjecture in Step 3 is huge, and we do not have a general theorem that enables us to automate the verification part in Step 4. While Theorem 3 enables us to automate the proof of the sufficient part of the assertion (6) for a class of formulas $\Phi$, we do not have a similar result for the necessary part - as we shall see below, Theorem 4 helps a lot here, but it does not provide an automated procedure. Nonetheless, computers play a crucial role in all steps, and in the following we report some of the theorems discovered using the above procedure.

### 5.1 The 0-1-0 Problem

This problem asks if a given rule is strongly equivalent to the empty set, thus can always be deleted from any program. We have the following experimental result:

**Lemma 2** *If a rule $r$ mentions at most three distinct atoms, then $r \simeq_{se} \emptyset$ iff*

$$(Hd_r \cup Ng_r) \cap Ps_r \neq \emptyset.$$

Using Theorem 4, we can show the following result:

**Lemma 3** *If there is a rule $r$ of the form (1) such that $r \simeq_{se} \emptyset$ and $(Hd_r \cup Ng_r) \cap Ps_r \neq \emptyset$ is not true, then there is such a rule that mentions at most three atoms.*

**Proof:** Suppose $r \simeq_{se} \emptyset$, $Hd_r \cap Ps_r = \emptyset$, and $Ps_r \cap Ng_r = \emptyset$. Suppose $L$ is the set of atoms in $r$, and $a, b, c$ are three new atoms. Let

$$f(p) = \begin{cases} a & p \in Hd_r \\ b & p \in Ps_r \\ c & otherwise \end{cases}$$

By Theorem 4, we also have $f(r) \simeq_{se} \emptyset$. By the construction of $f$, we also have $Hd_{f(r)} \cap Ps_{f(r)} = \emptyset$, and $Ps_{f(r)} \cap Ng_{f(r)} = \emptyset$, and that $f(r)$ mentions at most three distinct atoms. □

**Theorem 5 (The 0-1-0 problem)** *Lemma 2 holds in the general case, i.e. without any restriction on the number of atoms in $r$.*

**Proof:** We notice that the condition in Lemma 2, $(Hd_r \cup Ng_r) \cap Ps_r \neq \emptyset$, is equivalent to the following property

$$\exists x.(X_1(x) \vee Z_1(x)) \wedge Y_1(x)$$

being true on $[r]$. Thus the "if" part follows from Theorem 3 and Lemma 2. The "only if" part follows from Lemma 2 and Lemma 3. □

The "if" part of the theorem is already well-known, first proved by Osorio *et. al.* (2001). The "only if" part has also been proved recently by Inoue and Sakama (2004). While we did not discover anything new in this case, it is reassuring that the methodology works.

We notice here that there is no need to consider the 0-$n$-0 problem for $n > 1$, because for any $n$, $\{r_1, ..., r_n\}$ is strongly equivalent to $\emptyset$ iff for each $1 \leq i \leq n$, $\{r_i\}$ is strongly equivalent to $\emptyset$.

## 5.2 The 1-1-0 and the 0-1-1 Problems

The 1-1-0 problem asks if a rule can always be deleted in the presence of another rule, and the 0-1-1 problem asks if a rule can always be replaced by another one. We first solve the 1-1-0 problem, and the solution to the 0-1-1 problem will come as a corollary.

We have the following experimental result for the 1-1-0 problem:

**Lemma 4** *For any two rules $r_1$ and $r_2$ that mentions at most three atoms, $\{r_1, r_2\}$ and $\{r_1\}$ are strongly equivalent iff one of the following two conditions is true:*

1. *$r_2 \simeq_{se} \emptyset$.*

2. *$Ps_{r_1} \subseteq Ps_{r_2}$, $Ng_{r_1} \subseteq Ng_{r_2}$, and $Hd_{r_1} \subseteq Hd_{r_2} \cup Ng_{r_2}$.*

**Lemma 5** *If there are two rules $r_1$ and $r_2$ such that $\{r_1, r_2\} \simeq_{se} \{r_2\}$, but none of the two conditions in Lemma 4 hold, then there are two such rules that mention at most three atoms.*

**Proof:** Suppose there are two rules $r_1, r_2$ such that $\{r_1, r_2\} \simeq_{se} \{r_2\}$, and none of the two conditions in Lemma 4 hold. Let $L$ be the set of atoms in $r_1, r_2$.

Without loss of generality, suppose $a_1$ is an atom that makes the condition (2) in Lemma 4 false. If $Ps_{r_2} \setminus \{a_1\}$ is not empty, let $a_2$ be an atom in it. Let $L' = \{a_1, a_2, a_3\}$, where $a_3$ is a new atom, and $f$ be a function from $L$ to $L'$ as following:

$$f(a) = \begin{cases} a_1 & a = a_1 \\ a_2 & a \in Ps_{r_2} \setminus \{a_1\} \\ a_3 & otherwise \end{cases}$$

clearly, $f(r_1)$ and $f(r_2)$ mention at most three distinct atoms, and by Theorem 4,

$$\{f(r_1), f(r_2)\} \simeq_{se} f(r_1).$$

We show that none of the two conditions in Lemma 4 hold for $f(r_1)$ and $f(r_2)$ either.

We show first that $f(r_2) \not\simeq_{se} \emptyset$. By Theorem 5, we need to show that

$$S = Ps_{f(r_2)} \cap (Hd_{f(r_2)} \cup Ng_{f(r_2)})$$

is empty. If $a_1 \in S$, then by the construction of $f$, $a_1 \in Ps_{r_2} \cap (Hd_{r_2} \cup Ng_{r_2})$, a contradiction with the assumption that $r_2$ is not strongly equivalent to $\emptyset$. Similarly, if $a_2 \in S$, then by the construction of $f$, $a_2 \in Ps_{r_2} \cap (Hd_{r_2} \cup Ng_{r_2})$, a contradiction with the assumption that $r_2$ is not strongly equivalent to $\emptyset$. But then $a_3$ cannot be in $S$ as $a_3$ cannot be in $Ps_{f(r_2)}$. Thus $S$ must be empty.

We now show that it is not the case that $Ps_{f(r_1)} \subseteq Ps_{f(r_2)}$, $Ng_{f(r_1)} \subseteq Ng_{f(r_2)}$, and $Hd_{f(r_1)} \subseteq Hd_{f(r_2)} \cup Ng_{f(r_2)}$. By our assumption, $a_1$ is an atom that makes either $Ps_{r_1} \subseteq Ps_{r_2}$, $Ng_{r_1} \subseteq Ng_{r_2}$, or $Hd_{r_1} \subseteq Hd_{r_2} \cup Ng_{r_2}$ false. There are three cases here. Suppose $a_1$ makes $Ps_{r_1} \subseteq Ps_{r_2}$ false, i.e. $a_1 \in Ps_{r_1}$ but $a_1 \notin Ps_{r_2}$. Then by our construction of $f$, we also have that $a_1 \in Ps_{f(r_1)}$ but $a_1 \notin Ps_{f(r_2)}$. The other two cases are similar. $\square$

**Theorem 6 (The 1-1-0 problem)** *Lemma 4 holds in the general case, without any restriction on the number of atoms in $r_1$ and $r_2$.*

**Proof:** The condition in Lemma 4 is equivalent to the following property

$$[\exists x.(X_2(x) \vee Z_2(x)) \wedge Y_2(x)] \vee$$
$$\{[\forall x.Y_1(x) \supset Y_2(x)] \wedge [\forall x.Z_1(x) \supset Z_2(x)] \wedge [\forall x.X_1(x) \supset (X_2(x) \vee Z_2(x))]\}$$

being true on $[r_1, r_2]$. Thus the "if" part follows from Theorem 3 and Lemma 4, by noticing that the above property can be written as $\exists x \forall \vec{y}.Q$ as required by Theorem 3. The "only if" part follows from Lemma 4 and Lemma 5. $\square$

Thus if a rule $r_2$ cannot be deleted on its own but can be deleted in the presence of another rule $r_1$, then it must be the case that $r_2$ is redundant given $r_1$: if the body of $r_2$ is satisfied, then the body of $r_1$ is satisfied as well; furthermore, $r_2$ can entail no more than what can be entailed by $r_1$ ($Hd_{r_1} \subseteq Hd_{r_2} \cup Ng_{r_2}$).

Osorio et al. (2001) proved that $\{r_1, r_2\} \simeq_{se} r_1$ if either $Ps_{r_1} \cup Ng_{r_1} = \emptyset$ and $Hd_{r_1} \subseteq Ng_{r_2}$ or $Ps_{r_1} \subseteq Ps_{r_2}$, $Ng_{r_1} \subseteq Ng_{r_2}$, and $Hd_{r_1} \subseteq Hd_{r_2}$. More recently, Eiter et al. (2004) showed that $\{r_1, r_2\} \simeq_{se} r_1$ if $r_1$ s-implies $r_2$ (Wang & Zhou, 2005), i.e. if there exists a set $A \subseteq Ng_{r_2}$ such that $Hd_{r_1} \subseteq Hd_{r_2} \cup A$, $Ng_{r_1} \subseteq Ng_{r_2} \setminus A$, and $Ps_{r_1} \subseteq Ps_{r_2}$.

As one can see, these are all special cases of the "if" part of Theorem 6. Our result is actually more general. For instance, these special cases do not apply to

$$\{(c \leftarrow b, not\ c),\ (\leftarrow b, not\ c)\}$$

and

$$\{c \leftarrow b, not\ c\},$$

but one can easily show that these two sets are strongly equivalent using our theorem.

¿From our solution to the 1-1-0 problem, we can derive a solution to the 0-1-1 problem.

**Theorem 7 (The 0-1-1 problem)** *For any two rules $r_1$ and $r_2$, $r_1 \simeq_{se} r_2$ iff one of the following two conditions is true:*

1. $r_1 \simeq_{se} r_2 \simeq_{se} \emptyset$.

2. $Ps_{r_1} = Ps_{r_2}$, $Ng_{r_1} = Ng_{r_2}$, and $Hd_{r_1} \cup Ng_{r_1} = Hd_{r_2} \cup Ng_{r_2}$.

**Proof:** By Theorem 1, it is easy to see that $r_1 \simeq_{se} r_2$ iff $\{r_1, r_2\} \simeq_{se} r_1$ and $\{r_1, r_2\} \simeq_{se} r_2$.
□

Thus two rules $r_1$ and $r_2$ can always be interchanged if either both of them can be deleted (strongly equivalent to the empty set) or they have the same body, and the same consequences when the body is true. For instance, we have $\{a \leftarrow B, not\ a\} \simeq_{se} \{\leftarrow B, not\ a\}$ no matter what $B$ is, because the two rules have the same body, and when the body is true, the same consequence - a contradiction. As another example, we have

$$\{a; b \leftarrow not\ a\} \simeq_{se} \{b \leftarrow not\ a\},$$

because the two rules have the same body, and, when the body is true, the same consequence, $b$.

### 5.3 The 2-1-0, 0-2-1, and 0-2-2 Problems

The 2-1-0 problem asks if a rule can be deleted in the presence of another two rules, the 0-2-1 problem asks if two rules can be replaced by a single rule, and the 0-2-2 problem asks if two rules can be replaced by another two rules. Similar to the previous subsection, the solution to the 0-2-1 and 0-2-2 problems will follow from a solution to the 2-1-0 problem.

The experiment on the 2-1-0 problem was more difficult because as it turned out, we have to consider a language with six atoms in this case. In principle, given a language $L$, every subset of $L$ can be the $Hd$, $Ps$, or $Ng$ of a rule. Thus when the size of $L$ is six, there are in principle $(2^6)^3 - 1 = 262,143$ possible rules, and $262,143^3$ triples of them. However, we can cut down the numbers significantly with the results that we already have proved.

First, we only have to consider rules that do not have common elements in any of the two sets in $\{Hd, Ps, Ng\}$: if either $Hd$ and $Ps$ or $Ps$ and $Ng$ have a common element, then by Theorem 5, this rule can be deleted; if $Hd$ and $Ng$ have common elements, then according to Theorem 7, we obtain a strongly equivalent rule by deleting the common elements in $Hd$. In the following, we call such rules *canonical*, that is, a rule $r$ is canonical if

$$Hd_r \cap Ps_r = Hd_r \cap Ng_r = Ps_r \cap Ng_r = \emptyset.$$

Secondly, we do not have to consider isomorphic rules: if there is a one-to-one onto function from $L$ to $L$ that maps $\{r_1, r_2, r_3\}$ to $\{r'_1, r'_2, r'_3\}$, then these two sets of rules are essentially the same except for the names of atoms in them.

Thus by considering only canonical rules and using a certain normal form for triples of rules that avoids isomorphic rules, we ended up with roughly 120 million triples of rules to consider for verifying the following result, which took about 10 hours on a Solaris server consisting of 8 Sun Ultra-SPARC III 900Mhz CPUs with 8GB RAM.

For more details on the experiment on 2-1-0 problem, please refer to (Chen, Lin, & Li, 2005).

**Lemma 6** *For any three canonical rules $r_1$, $r_2$ and $r_3$ that mention at most six atoms, $\{r_1, r_2, r_3\} \simeq_{se} \{r_1, r_2\}$ iff one of the following three conditions is true:*

1. *$\{r_1, r_3\} \simeq_{se} r_1$.*

2. *$\{r_2, r_3\} \simeq_{se} r_2$.*

3. *There is an atom $p$ such that:*

    *3.1 $p \in (Ps_{r_1} \cup Ps_{r_2}) \cap (Hd_{r_1} \cup Hd_{r_2} \cup Ng_{r_1} \cup Ng_{r_2})$*

    *3.2 $Hd_{r_i} \setminus \{p\} \subseteq Hd_{r_3} \cup Ng_{r_3}$ and $Ps_{r_i} \setminus \{p\} \subseteq Ps_{r_3}$ and $Ng_{r_i} \setminus \{p\} \subseteq Ng_{r_3}$, where $i = 1, 2$*

    *3.3 If $p \in Ps_{r_1} \cap Ng_{r_2}$, then $Hd_{r_1} \cap Hd_{r_3} = \emptyset$*

    *3.4 If $p \in Ps_{r_2} \cap Ng_{r_1}$, then $Hd_{r_2} \cap Hd_{r_3} = \emptyset$*

The following lemma is the reason why we need to consider a language with six atoms for this problem.

**Lemma 7** *If there are three canonical rules $r_1, r_2$ and $r_3$ such that $\{r_1, r_2, r_3\} \simeq_{se} \{r_1, r_2\}$, but none of the three conditions in Lemma 6 hold, then there are three such rules that mention at most six atoms.*

**Proof:** The proof of this lemma is tedious as we have to consider several cases. Consider the following statements about any three canonical rules $r_1, r_2, r_3$:

(I) $\{r_1, r_2, r_3\} \simeq_{se} \{r_1, r_2\}$.

(II) $\{r_1, r_3\} \not\simeq_{se} \{r_1\}$, i.e. $Ps_{r_1} \not\subseteq Ps_{r_3}$ or $Ng_{r_1} \not\subseteq Ng_{r_3}$ or $Hd_{r_1} \cup Ng_{r_1} \not\subseteq Hd_{r_3} \cup Ng_{r_3}$

(III) $\{r_2, r_3\} \not\simeq_{se} \{r_2\}$, i.e. $Ps_{r_2} \not\subseteq Ps_{r_3}$ or $Ng_{r_2} \not\subseteq Ng_{r_3}$ or $Hd_{r_2} \cup Ng_{r_2} \not\subseteq Hd_{r_3} \cup Ng_{r_3}$

(IV) $(Ps_{r_1} \cup Ps_{r_2}) \cap (Hd_{r_1} \cup Hd_{r_2} \cup Ng_{r_1} \cup Ng_{r_2}) = \emptyset$

(V) There is an atom $p$ in the set $(Ps_{r_1} \cup Ps_{r_2}) \cap (Hd_{r_1} \cup Hd_{r_2} \cup Ng_{r_1} \cup Ng_{r_2})$, and another different atom $q$ such that one of the following three conditions is true:

  1. $q \in Hd_{r_1} \cup Ng_{r_1}$ and $q \notin Hd_{r_3} \cup Ng_{r_3}$.
  2. $q \in Ps_{r_1}$ and $q \notin Ps_{r_3}$.
  3. $q \in Ng_{r_1}$ and $q \notin Ng_{r_3}$.

  Notice that this is the negation of condition (3.2) in Lemma 6.

(VI) $Hd_{r_1} \cap Hd_{r_3} \not\subseteq Ng_{r_3}$, and there is an atom $p \in Ps_{r_1} \cap Ng_{r_2}$ such that for $i = 1, 2$, $Hd_{r_i} \setminus \{p\} \subseteq Hd_{r_3} \cup Ng_{r_3}$, $Ps_{r_i} \setminus \{p\} \subseteq Ps_{r_3}$, and $Ng_{r_i} \setminus \{p\} \subseteq Ng_{r_3}$.

Since $r_1$ and $r_2$ are symmetric in the conditions in Lemma 6, to prove this lemma, we need only to prove the following three assertions:

(a) If there are three canonical rules $r_1, r_2, r_3$ which satisfy (I)-(IV), then there are three canonical rules $r'_1, r'_2, r'_3$ which mention at most six atoms, and satisfy (I)-(IV) as well.

(b) If there are three canonical rules $r_1, r_2, r_3$ which satisfy (I)-(III)(V), then there are three canonical rules $r'_1, r'_2, r'_3$ which mention at most six atoms, and satisfy (I)-(III)(V) as well.

(c) If there are three canonical rules $r_1, r_2, r_3$ which satisfy (I)-(III)(VI), then there are three canonical rules $r'_1, r'_2, r'_3$ which mention at most six atoms, and satisfy (I)-(III)(VI) as well.

We now prove the above three assertions one by one.

(a) Let $a_1, a_2$ be two atoms that make (II) and (III) true. If $(Ps_{r_3} \cap (Ps_{r_1} \cup Ps_{r_2})) \setminus \{a_1, a_2\}$ is not empty, let $a_3$ be an atom in it. If $Ps_{r_3} \setminus (Ps_{r_1} \cup Ps_{r_2} \cup \{a_1, a_2\})$ is not empty, let $a_4$ be an atom in it. If $(Ps_{r_1} \cup Ps_{r_2}) \setminus (Ps_{r_3} \cup \{a_1, a_2\})$ is not empty, let $a_5$ be an atom in it. Finally let $a_6$ be a new atom different from $a_1$ to $a_5$, and $L' = \{a_1, a_2, a_3, a_4, a_5, a_6\}$. Let $f$ be a function from $L$ to $L'$ defined as following:

$$f(a) = \begin{cases} a_1 & a = a_1 \\ a_2 & a = a_2 \\ a_3 & a \in (Ps_{r_3} \cap (Ps_{r_1} \cup Ps_{r_2})) \setminus \{a_1, a_2\} \\ a_4 & a \in Ps_{r_3} \setminus (Ps_{r_1} \cup Ps_{r_2} \cup \{a_1, a_2\}) \\ a_5 & a \in (Ps_{r_1} \cup Ps_{r_2}) \setminus (Ps_{r_3} \cup \{a_1, a_2\}) \\ a_6 & otherwise \end{cases}$$

For each $1 \leq i \leq 3$, let $r'_i$ be as follows:

$$Ps_{r'_i} = Ps_{f(r_i)}, Ng_{r'_i} = Ng_{f(r_i)}, Hd_{r'_i} = Hd_{f(r_i)} \setminus Ng_{f(r_i)}. \tag{12}$$

We have that

- For each $1 \leq i \leq 3$, $r'_i$ is a canonical rule, and $r'_i \simeq_{se} f(r_i)$. For this, we only need to show $f(r_i) \not\simeq_{se} \emptyset$ for each $1 \leq i \leq 3$. To see this, notice that from the definition of $f$, atoms other than $a_1$ and $a_2$ in $Ps_{r_3}$ are mapped to $\{a_3, a_4\}$, and atoms other than $a_1$ and $a_2$ in $Hd_{r_3} \cup Ng_{r_3}$ are mapped to $\{a_5, a_6\}$. Thus $Ps_{f(r_3)} \cap (Hd_{f(r_3)} \cup Ng_{f(r_3)}) = \emptyset$. By Theorem 5, $f(r_3) \not\simeq_{se} \emptyset$. Now $f(r_1) \not\simeq_{se} \emptyset$ and $f(r_2) \not\simeq_{se} \emptyset$, because (II) and (III) hold for $f(r_1), f(r_2), f(r_3)$ by definition of $f$.
- (I) holds for $r'_1, r'_2, r'_3$. This is because by Theorem 4,

$$\{f(r_1), f(r_2), f(r_3)\} \not\simeq_{se} \{f(r_1), f(r_2)\},$$

and for each $1 \leq i \leq 3$, $r'_i \simeq_{se} f(r_i)$.

- (II) and (III) hold for $r'_1, r'_2, r'_3$. As we mentioned, from the definition of $f$, (II) and (III) hold for $f(r_1), f(r_2), f(r_3)$.

- (IV) holds for $r'_1, r'_2, r'_3$. Again, we need only to show that (IV) holds for $f(r_1), f(r_2), f(r_3)$. To see this, notice that atoms other than $a_1$ and $a_2$ in $Ps_{r_1} \cup Ps_{r_2}$ are mapped to $\{a_3, a_5\}$, and atoms other than $a_1$ and $a_2$ in $Hd_{r_1} \cup Hd_{r_2} \cup Ng_{r_1} \cup Ng_{r_2}$ are mapped to $\{a_4, a_6\}$.

(b) Again let $a_1, a_2$ be two atoms that make (II) and (III) true. Let $p, q$ be the two witness atoms in (V). If $Pos(r_3) \setminus \{a_1, a_2, p, q\}$ is not empty, let $a_3$ be an atom in it. Let $a_4$ be a new atom, and $L' = \{a_1, a_2, a_3, a_4, p, q\}$. Define $f$ as follows:

$$f(a) = \begin{cases} a_1 & a = a_1 \\ a_2 & a = a_2 \\ p & a = p \\ q & a = q \\ a_3 & a \in Ps_{r_3} \setminus \{a_1, a_2, p, q\} \\ a_4 & otherwise \end{cases}$$

Define $r'_i$ by (12) as well for each $1 \leq i \leq 3$.

- For each $1 \leq i \leq 3$, $r'_i$ is a canonical rule, and $r'_i \simeq_{se} f(r_i)$. This can be seen in the same way as for (a) above.
- By Theorem 4, $\{f(r_1), f(r_2), f(r_3)\} \simeq_{se} \{f(r_1), f(r_2)\}$, and thus

$$\{r'_1, r'_2, r'_3\} \simeq_{se} \{r'_1, r'_2\}.$$

  So (I) holds for $r'_1, r'_2, r'_3$.
- From definition of $f$, (II) and (III) hold for $f(r_1), f(r_2), f(r_3)$, and thus they hold for $r'_1, r'_2, r'_3$ as well.
- Again from the definition of $f$, (V) holds for $f(r_1), f(r_2), f(r_3)$: there is an atom $p$ in the set $(Ps_{f(r_1)} \cup Ps_{f(r_2)}) \cap (Hd_{f(r_1)} \cup Hd_{f(r_2)} \cup Ng_{f(r_1)} \cup Ng_{f(r_2)})$, and another different atom $q$ such that one of the following three conditions is true:
  1. $q \in Hd_{f(r_1)} \cup Ng_{f(r_1)}$ and $q \notin Hd_{f(r_3)} \cup Ng_{f(r_3)}$.
  2. $q \in Ps_{f(r_1)}$ and $q \notin Ps_{f(r_3)}$.
  3. $q \in Ng_{f(r_1)}$ and $q \notin Ng_{f(r_3)}$.
  
  (V) holds for $r'_1, r'_2, r'_3$ as well because for each $1 \leq i \leq 3$,
  
  $$Ps_{r'_i} = Ps_{f(r_i)}, Ng_{r'_i} = Ng_{f(r_i)}, Hd_{r'_i} \cup Ng_{r'_i} = Hd_{f(r_i)} \cup Ng_{f(r_i)}.$$

(c) Let $a_1, a_2$ be two atoms that make (II) and (III) true. Let $p$ be the witness atom in (VI), and let $q \in Hd_{r_1} \cap Hd_{r_3}$ but $q \notin Ng_{r_3}$. If $Pos(r_3) \setminus \{a_1, a_2, p, q\}$ is not empty, let $a_3$ be an atom in it. Let $a_4$ is a new atom, and Let $L' = \{a_1, a_2, a_3, a_4, p, q\}$, Define $f$ as follows:

$$f(a) = \begin{cases} a_1 & a = a_1 \\ a_2 & a = a_2 \\ p & a = p \\ q & a = q \\ a_3 & a \in Ps_{r_3} \setminus \{a_1, a_2, p, q\} \\ a_4 & otherwise \end{cases}$$

Again define $r'_i$ by (12) as well for each $1 \leq i \leq 3$.

- For each $1 \leq i \leq 3$, $r'_i$ is a canonical rule, and $r'_i \simeq_{se} f(r_i)$. This can be seen in the same way as for (a) above.

445

- Again by Theorem 4, $\{f(r_1), f(r_2), f(r_3)\} \simeq_{se} \{f(r_1), f(r_2)\}$, and thus

$$\{r'_1, r'_2, r'_3\} \simeq_{se} \{r'_1, r'_2\}.$$

So (I) holds for $r'_1, r'_2, r'_3$.
- Again from definition of $f$, (II) and (III) hold for $f(r_1), f(r_2), f(r_3)$, thus they hold for $r'_1, r'_2, r'_3$ as well.
- By the definition of $f$, (VI) holds for $f(r_1), f(r_2), f(r_3)$: $Hd_{f(r_1)} \cap Hd_{f(r_3)} \nsubseteq Ng_{f(r_3)}$, and there is an atom $p \in Ps_{f(r_1)} \cap Ng_{f(r_2)}$ such that for $i = 1, 2$, $Hd_{f(r_i)} \setminus \{p\} \subseteq Hd_{f(r_3)} \cup Ng_{f(r_3)}$, $Ps_{f(r_i)} \setminus \{p\} \subseteq Ps_{f(r_3)}$, and $Ng_{f(r_i)} \setminus \{p\} \subseteq Ng_{f(r_3)}$. (VI) holds for $r'_1, r'_2, r'_3$ as well because

$$Ps_{r'_i} = Ps_{f(r_i)}, Ng_{r'_i} = Ng_{f(r_i)}, Hd_{r'_i} \subseteq Hd_{f(r_i)}.$$

□

**Theorem 8 (The 2-1-0 problem)** *Lemma 6 holds in the general case, without any restriction on the number of atoms in $r_1, r_2, r_3$.*

**Proof:** The assertion that $r_1$, $r_2$, and $r_3$ are canonical rules and satisfy one of the three conditions in Lemma 6 is equivalent to the following property

$$[\forall x.((\neg(X_1(x) \wedge Y_1(x))) \wedge (\neg(X_1(x) \wedge Z_1(x))) \wedge (\neg(Y_1(x) \wedge Z_1(x))))] \wedge$$
$$[\forall x.((\neg(X_2(x) \wedge Y_2(x))) \wedge (\neg(X_2(x) \wedge Z_2(x))) \wedge (\neg(Y_2(x) \wedge Z_2(x))))] \wedge$$
$$[\forall x.((\neg(X_3(x) \wedge Y_3(x))) \wedge (\neg(X_3(x) \wedge Z_3(x))) \wedge (\neg(Y_3(x) \wedge Z_3(x))))] \wedge$$
$$\{[(\forall x.Y_1(x) \supset Y_3(x)) \wedge (\forall x.Z_1(x) \supset Z_3(x)) \wedge (\forall x.X_1(x) \supset (X_3(x) \vee Z_3(x)))] \vee$$
$$[(\forall x.Y_2(x) \supset Y_3(x)) \wedge (\forall x.Z_2(x) \supset Z_3(x)) \wedge (\forall x.X_2(x) \supset (X_3(x) \vee Z_3(x)))] \vee$$
$$[\exists x.CON1(x) \wedge CON2(x) \wedge CON3(x) \wedge CON4(x)]\}$$

being true on $[r_1, r_2, r_3]$, where $CON1(x)$ stands for

$$(Y_1(x) \vee Y_2(x)) \wedge (X_1(x) \vee X_2(x) \vee Z_1(x) \vee Z_2(x))$$

$CON2(x)$ for

$$\forall y.(x \neq y) \supset [(X_1(y) \supset (X_3(y) \vee Z_3(y))) \wedge (Y_1(y) \supset Y_3(y)) \wedge (Z_1(y) \supset Z_3(y)) \wedge$$
$$(X_2(y) \supset (X_3(y) \vee Z_3(y))) \wedge (Y_2(y) \supset Y_3(y)) \wedge (Z_2(y) \supset Z_3(y))]$$

$CON3(x)$ for

$$Y_1(x) \wedge Z_2(x) \supset \forall y.(\neg(X_1(y) \wedge X_3(y))),$$

and $CON4(x)$ for

$$Y_2(x) \wedge Z_1(x) \supset \forall y.(\neg(X_2(y) \wedge X_3(y))).$$

Thus the "if" part follows from Theorem 3 and Lemma 6, by noticing that the above property can be written as $\exists x \forall \vec{y}.Q$ as required by Theorem 3. The "only if" part follows from Lemma 6 and Lemma 7. □

The conditions in Lemma 6 (Theorem 8) are rather complex, and the reason why it is difficult to automate Step 3 of the procedure at the beginning of the section. These conditions capture all possible cases when $r_3$ is "subsumed" by $r_1$ and $r_2$, and are difficult to describe concisely by words. We give some examples.

Consider the following three rules:

$$r_1 : (a_2 \leftarrow a_1)$$
$$r_2 : (a_3 \leftarrow not\ a_1)$$
$$r_3 : (a_3 \leftarrow not\ a_2).$$

We have that $\{r_1, r_2, r_3\} \simeq_{se} \{r_1, r_2\}$ because the condition (4) in Lemma 6 holds.

However, if we change $r_3$ into $r_3' : a_2 \leftarrow not\ a_3$, then $P_1 = \{r_1, r_2, r_3'\}$ and $P_2 = \{r_1, r_2\}$ are not strongly equivalent: one could check that condition (4.3) in Lemma 6 does not hold, and indeed, while $P_2 \cup \{a_1 \leftarrow a_2\}$ has a unique answer set $\{a_3\}$, $P_1 \cup \{a_1 \leftarrow a_2\}$ has two answer sets $\{a_3\}$ and $\{a_1, a_2\}$.

It is also easy to show by Theorem 8 that $a_3 \leftarrow not\ a_2$ is "subsumed" by

$$\{(a_1; a_2; a_3 \leftarrow), (a_2; a_3 \leftarrow a_1)\},$$

and $a_2; a_3 \leftarrow$ is "subsumed" by

$$\{(a_2 \leftarrow a_1), (a_3 \leftarrow not\ a_1)\}.$$

With the results that we have, the following theorem will yield a solution to the 0-2-1 problem.

**Theorem 9 (the 0-2-1 problem)** *For any three rules $r_1$, $r_2$ and $r_3$, $\{r_1, r_2\}$ and $\{r_3\}$ are strongly equivalent iff the following three conditions are true:*

1. $\{r_1, r_2, r_3\} \simeq_{se} \{r_1, r_2\}$.
2. $\{r_1, r_3\} \simeq_{se} \{r_3\}$.
3. $\{r_2, r_3\} \simeq_{se} \{r_3\}$.

For example, We have

$$\{(a_2 \leftarrow a_1, not\ a_3), (a_1; a_2 \leftarrow not\ a_3)\} \simeq_{se} \{a_2 \leftarrow not\ a_3\}.$$

While we have

$$\{(\leftarrow a_2, a_3), (\leftarrow a_3, not\ a_2)\} \simeq_{se} \{\leftarrow a_3\},$$

we have

$$\{(a_1 \leftarrow a_2, a_3), (a_1 \leftarrow a_3, not\ a_2)\} \not\simeq_{se} \{a_1 \leftarrow a_3\}.$$

Similarly, we have the following theorem

**Theorem 10 (the 0-2-2 problem)** *For any four rules $r_1$, $r_2$, $r_3$, $r_4$, $\{r_1, r_2\}$ and $\{r_3, r_4\}$ are strongly equivalent iff the following four conditions are true:*

1. $\{r_1, r_2, r_3\} \simeq_{se} \{r_1, r_2\}$.

2. $\{r_1, r_2, r_4\} \simeq_{se} \{r_1, r_2\}$.

3. $\{r_3, r_4, r_1\} \simeq_{se} \{r_3, r_4\}$.

4. $\{r_3, r_4, r_2\} \simeq_{se} \{r_3, r_4\}$.

## 6. Program Simplification

We have mentioned that one possible use of the notion of strongly equivalent logic programs is in simplifying logic programs: if $P \simeq_{se} Q$, and that $Q$ is "simpler" than $P$, we can then replace $P$ in any program that contains it by $Q$.

Most answer set programming systems perform some program simplifications. However, only Smodels (Niemelä et al., 2000) has a stand-alone front-end called lparse that can be used to ground and simplify a given logic program. It seems that lparse simplifies a grounded logic program by computing first its well-founded model. It does not, however, perform any program simplification using the notion of strong equivalence. For instance, lparse-1.0.13, the current version of lparse, did nothing to the following set of rules:
$\{(a \leftarrow not\, b), (b \leftarrow not\, a), (a \leftarrow a)\}$. Nor does it replace the first rule in the following program $\{(a \leftarrow not\, a), (a \leftarrow not\, b), (b \leftarrow not\, a)\}$ by the constraint $\leftarrow not\, a$.

It is unlikely that anyone would be intentionally writing rules like $a \leftarrow a$ or $b \leftarrow a, not\, a$. But these type of rules can arise as a result of grounding some rules with variables. For instance, the following is a typical recursive rule used in logic programming encoding of the Hamiltonian Circuit problem (Niemelä, 1999; Marek & Truszczynski, 1999):

$$reached(X) \leftarrow arc(Y, X), hc(Y, X), reached(Y).$$

When instantiated on a graph with cyclic arcs like $arc(a, a)$, this rule generates cyclic rules of the form $reached(X) \leftarrow hc(X, X), reached(X)$. Unless deleted explicitly, these rules will slow down many systems, especially those based on SAT. For instance, none of the graphs tested using ASSAT have self-cycles consisting of an arc from a node to itself (Lin & Zhao, 2004). If these cycles are included, ASSAT would run significantly longer.

It is thus useful to consider using the results that we have here for program simplification. Indeed, transformation rules such as deleting those that contain common elements in their heads and positive bodies have been proposed (Brass & Dix, 1999), and studied from the perspective of strong equivalence (Osorio et al., 2001; Eiter et al., 2004). Our results add new such transformation rules. For instance, by Theorem 7, we can delete those elements in the head of a rule that also appear in the negation-as-failure part of the rule. Theorems 6, 8, and 9 can also be used to define some new transformation rules.

## 7. Concluding Remarks and Future Work

Donald Knuth, in his Forward to (Petkovsek et al., 1996), said

> "Science is what we understand well enough to explain to a computer. Art is everything else we do. ...Science advances whenever an Art becomes a Science.

And the state of the Art advances too, because people always leap into new territory once they have understood more about the old."

We hope that with this work, we are one step closer to making discovering classes of strongly equivalent logic programs a Science.

We have mentioned that the methodology used in this paper is similar to that in (Lin, 2004). In both cases, plausible conjectures are generated by testing them in domains of small sizes, and general theorems are proved to aid the verification of these conjectures in the general case. However, while plausible conjectures are generated automatically in (Lin, 2004), they are done manually here. While the verifications of most conjectures in (Lin, 2004) are done automatically as well, they are done only semi-automatically here. Overcoming these two weaknesses is the focus of our future work. Specifically, we would like to make Step 3 of the procedure in Section 5 automatic, and prove a theorem similar to Theorem 3 to automate the proofs of the "only if" parts of theorems like Theorems 5 - 8, in the same way that Theorem 3 makes the proofs of the "if" parts of these theorems automatic. This way, we would be able to discover more interesting theorems in this area, and more easily!

## Acknowledgments

An extended abstract of this paper appeared in *Proceedings of IJCAI'2005*. We thank Yan Zhang for his comments on an earlier version of the paper. We also thank the anonymous reviewers for their useful comments, especially one of them for pointing out an error in Lemma 4 in an earlier version of the paper. This work was supported in part by the Research Grants Council of Hong Kong under Competitive Earmarked Research Grant HKUST6170/04E. Part of the second author's work was done when he was a student at Sun Yat-Sen University, Guangzhou, China, and a visiting scholar in Department of Computer Science and Engineering, Hong Kong University of Science and Technology, Hong Kong.